# Early Blindness Detection Based on Retinal Images Using Ensemble Learning

Niloy Sikder
*Computer Science & Engineering Discipline*
Khulna University
Khulna, Bangladesh
niloysikder333@gmail.com

Md. Sanaullah Chowdhury
*Electronics & Communication Engineering Discipline*
Khulna University
Khulna, Bangladesh
sanaullahashfat@gmail.com

Abu Shamim Mohammad Arif
*Computer Science & Engineering Discipline*
Khulna University
Khulna, Bangladesh
shamimarif@yahoo.com

Abdullah-Al Nahid
*Electronics & Communication Engineering Discipline*
Khulna University
Khulna, Bangladesh
nahid.ece.ku@gmail.com

*Abstract*—Diabetic retinopathy (DR) is the primary cause of vision loss among grown-up people around the world. In four out of five cases having diabetes for a prolonged period leads to DR. If detected early, more than 90% of the new DR occurrences can be prevented from turning into blindness through proper treatment. Despite having multiple treatment procedures available that are well-capable to deal with DR, the negligence and failure of early detection cost most of the DR patients their precious eyesight. The recent developments in the field of Digital Image Processing (DIP) and Machine Learning (ML) have paved the way to use machines in this regard. The contemporary technologies allow us to develop devices capable of automatically detecting the condition of a person's eyes based on their retinal images. However, in practice, several factors hinder the quality of the captured images and impede the detection outcome. In this study, a novel early blind detection method has been proposed based on the color information extracted from retinal images using an ensemble learning algorithm. The method has been tested on a set of retinal images collected from people living in the rural areas of South Asia, which resulted in a 91% classification accuracy.

## I. INTRODUCTION

Diabetes Mellitus (DM), widely known as diabetes, refers to a range of metabolic disorders that occur due to having an elevated level of blood sugar caused by inadequate production of insulin for an extended period. DM has become a serious health issue all around the world. According to the reports of the International Diabetes Federation, in 2015, 415 million people (worldwide) had some level of diabetes; and the number will rise to an overwhelming 552 million by 2030 [1]. Broadly, diabetes can be categorized into four types – Type 1, Type 2, Genetic deficits, and Gestational diabetes [2]. Diabetes causes numerous complications in various parts of the body, including the heart, kidney, nervous system, and eyes. Diabetes affects these organs by damaging their internal blood vessels and obstructing the flow of blood. Issues encountered by the retinas due to diabetes are collectively known as Diabetic Retinopathy (DR). Although there are other factors capable of damaging the retina, DR is the leading cause of blindness in adults. Globally, more than half of the blindness cases are caused by DR, and 56% of the people who lose eyesight because of DR belong to the Asia-Pacific region [3]. A 2013 study shows that in Bangladesh, the prevalence rate of retinal damage among diabetic patients and patients having impaired glucose regulation was 21.6% and 13%, respectively [4]. Also, the latest studies suggest that the situation is getting worse every year all over the world.

Diabetes causes retinopathy by vandalizing the blood vessels within the retinal tissue, primarily causing them to leak blood, fluids, and lipids inside the macula; and then, blocking them off completely. The person, as a result, experiences blurred vision, impaired color vision, floaters, poor night vision, and finally, total blindness. However, as frightening as the effects are, we have several effective treatments available to deal with DR, which are useful if the condition is detected early. But most people do not get their eyes checked regularly, and hence, remain utterly unaware of the condition before the symptoms become apparent. Using machines to fast-track the detection procedure and making the technology available to the common people so that they can easily determine the state of their eyes would be a viable solution to this problem, which has been the ultimate goal of developing Machine Learning (ML) models for DR.

The use of ML in blindness detection is not new. However, the approaches primarily differ in terms of the retinal image processing methods, types of extracted features, and the utilized ML algorithms. Since the detection is based on images, deep learning methods are mostly used in these procedures. Notably, the Convolutional Neural Network (CNN) and its variants are very popular and proven to be effective as well. However, in this study, we used an ensemble learning algorithm called the Extremely Randomized Tree (popularly known as ExtraTree or ET) classifier, which follows a simple working principle and provides quick and accurate classification outcomes to detect the level of blindness. Approaches similar to this (non-deep learning) are less common, but not rare. For instance, in 2013, [5] proposed a method for microaneurysms classification using Gaussian Mixture Model (GMM) and support vector machine (SVM). In 2014, [6] described a method for binary DR detection involving multiple decision tree-based classifiers. Later that year, [7] outlined a method incorporating CNN and Random Forest (RF) for retinal blood vessel segmentation. In 2017, [8] proposed a method for DR assessment using a fuzzy RF and dominance-based rough set balanced rule ensemble. The proposed method differs from these procedures in terms of the source of the retinal images, the employed image processing techniques, and type of features extracted from the images.

The rest of the paper is organized as follows. Section II contains the workflow diagram and discussions on each step of the methodology. Section III presents the experimental outcomes of the study, along with brief discussions. Finally, Section IV concludes the paper with an overview of the study and points out some scopes for further research.

## II. METHODOLOGY

The workflow of the proposed DR classification model is charted in Fig. 1. The whole procedure is divided into four segments, which are performed sequentially. The upcoming paragraphs contain details on each step of the methodology.

## A. Retinal Image Collection

In this study, the retinal images of the Asia Pacific Tele-Ophthalmology Society 2019 Blindness Detection (APTOS 2019 BD) dataset were used. APTOS 2019 BD is an open (Kaggle) competition [9]. The associated dataset contains 3,662 retinal images collected from numerous subjects living in the rural areas of India. The images were put together by Aravind Eye Hospital, India, with the purpose of building an ML model to detect blindness autonomously without medical screening. These Red-Green-Blue (RGB) images were captured using Fundus photography. The samples were then labeled by trained doctors who categorized the extent of blindness of a person into five different levels – no DR, mild DR, moderate DR, severe DR, and proliferative DR. However, according to the website, participates should expect noise in both images and their corresponding class labels, and images may contain artifacts and be out of focus. The images were captured in various conditions and environments over an extended period. Table I provides a sample image of each class of the APTOS 2019 BD dataset along with the labels assigned to these classes to refer to them afterward.

TABLE I.  SAMPLE IMAGES OF VARIOUS DR CLASSES

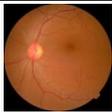

| No DR | Mild DR | Moderate DR | Severe DR | Proliferative DR |
|---|---|---|---|---|
| **No** | **Mi** | **Mo** | **Se** | **Pr** |

## B. Excluding Noisy Images

To detect and single out the noisy samples (images) present in the dataset, all 3,662 images were manually examined. The samples were excluded from further processing if they were out of focus, overexposed, underexposed, or containing artifacts. This elaborate investigation resulted in an exclusion of 602 noisy images. Fig. 2 displays some of the images excluded from this study. Also, Table II provides details on how the contents of the dataset had changed after this step.

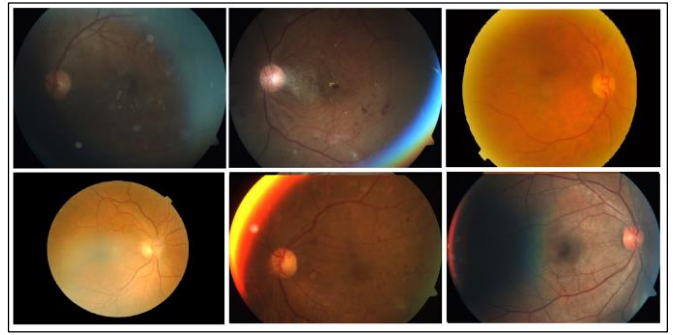

Fig. 2. A few of the images excluded from this study.

## C. Cropping Images to Eliminate Unwanted Information

Images of the APTOS 2019 BD dataset were not captured in the same condition using the same equipment. As a result, the resultant images contain different portions of the retina in different alignments. Some of the images contain the whole retina (Fig. 3(a)), but most of them miss the top and bottom segments (Fig. 3(c)). However, for blindness detection, the features of the reddish retina are required, not the black border around it. So, a crop operation was performed to select and cut-out the most significant part of the retinal and eliminate as much unnecessary black border as possible. Since our model is dependent on the color information of the images and in some cases, part of the retina is being excluded (Fig. 3(d)), this step holds great importance. The algorithm we developed to perform the crop operation is presented in *Algorithm-I*. To show the output of *Algorithm-I*, two sample images and their cropped area have been displayed in Fig. 3.

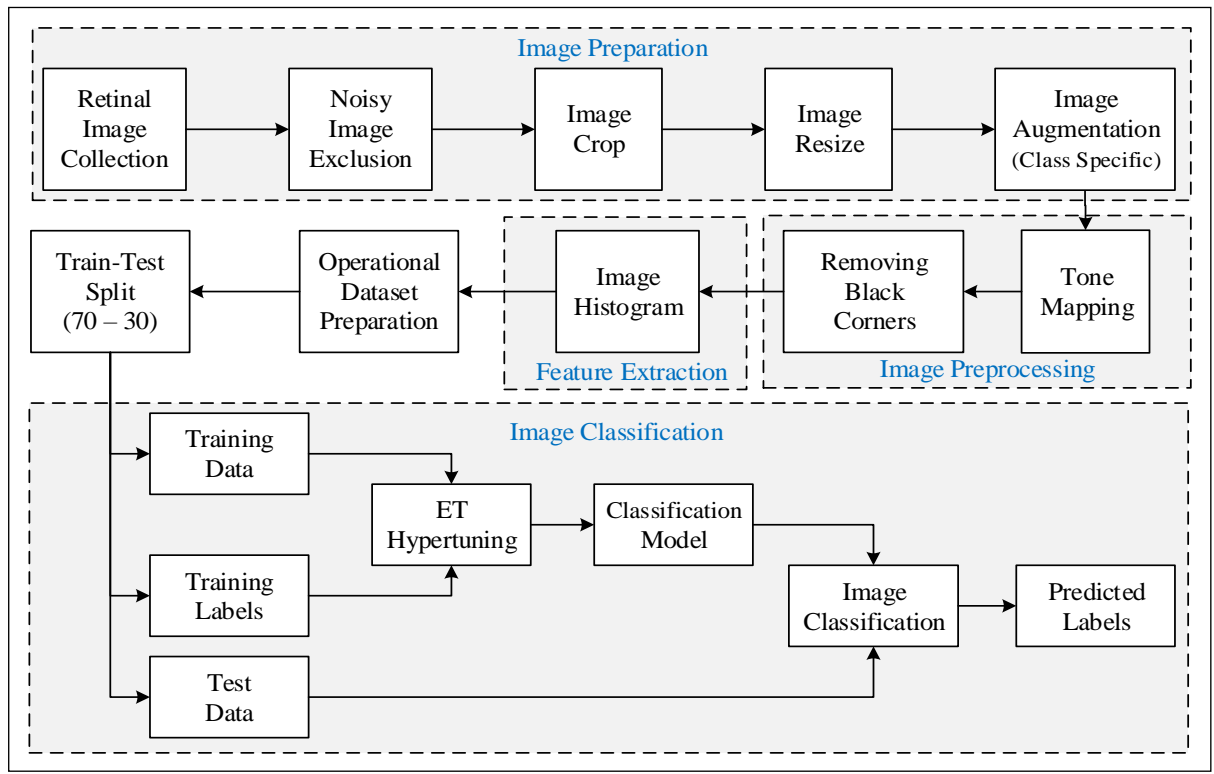

Fig. 1. Proposed methodology for retinal image classification.

| Algorithm-I: | | |
|---|---|---|
| Input | : | A retinal image, $I$ |
| Output | : | A cropped square-shaped image, $I'$ |
| Step-1 | : | Determine the height ($h$) and the width ($w$) of the image (in pixel). |
| Step-2 | : | Determine the center pixel:<br>$C_h = floor(h/2)$<br>$C_w = floor(w/2)$ |
| Step-3 | : | Determine the first non-black pixel directly above the center pixel at the top:<br>**for** $i = 1$ to $C_h$<br>  **if** the $(i, C_w)$ pixel is not zero in all the three channels (r, g, b)<br>    store the value of $i$ in $P$<br>    go to Step-4<br>**end for** |
| Step-4 | : | Determine the radius of the retina:<br>$R = C_h - P$ |
| Step-5 | : | Determine the boundaries for cropping:<br>$X_1 = (C_h - R)$, $X_2 = (C_h + R)$,<br>$Y_1 = (C_w - R)$, $Y_2 = (C_w + R)$ |
| Step-6 | : | Crop the image:<br>**for** $i = 1$ to $3$<br>  $I'(:,:,i) = I(X_1{:}X_2, Y_1{:}Y_2, i)$<br>**end for** |
| Step-7 | : | return $I'$ |

### D. Resizing Images

The collected images also vary in size and dimension. The dataset contains images ranging from $474 \times 358$ pixels to $3388 \times 2588$ pixels in width and height, respectively. However, to make them comparable, they have to be equal in size. That is why, in this step, an image resize operation is required. The proposed method is not bounded to any fixed image size or height-to-width ratio, as long as it is uniform.

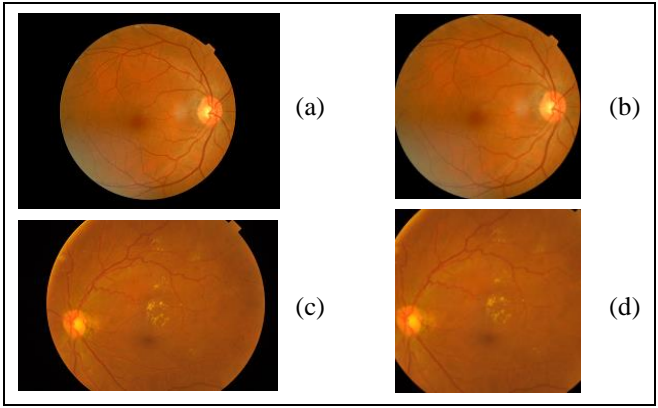

Fig. 3. (a) A sample image where the retinal has been captured entirely, (b) the targeted portion of (a), (c) a sample image where the retinal has been captured partially, (d) the targeted portion of (c).

### E. Image Augmentation

At this stage, we had 3,062 square-shaped retinal images. A quick look at Table II would reveal that even though collectively they belong to five different classes, the number of samples in each class vary significantly, which leads to an imbalanced dataset. Continuing the procedure with this sample ratio among the classes might lead to biasing in the classification stage. This issue seemed hard to overcome altogether since the provided dataset was imbalanced almost to the same extent. However, the images that belong to the inferior classes can be augmented to increase the number of samples of those classes, and in turn, roughly balance the ratio. In this study, we augmented the images that belong to the "Mi", "Se", and "Pr" classes in three different angles to acquire four times larger sample sizes than the original ones. Fig. 4 shows the augmented forms of a sample image. Table II shows the resultant sample ratio after this step.

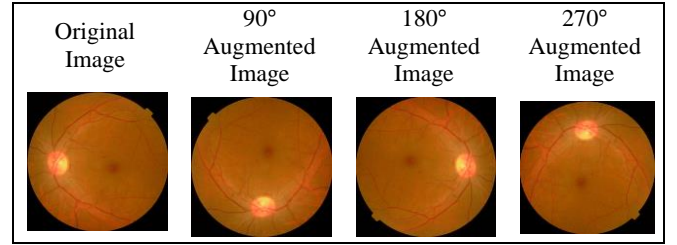

Fig. 4. Augmentation of a sample image at different conditions.

TABLE II. CLASS SAMPLE RATIO AT DIFFERENT STAGES

| Class | Number/Percentage of Images | | | | | |
|---|---|---|---|---|---|---|
| | APTOS Dataset | | After Image Exclusion | | After Image Augmentation | |
| No | 1805 | 49.3% | 1410 | 46.08% | 1410 | 26.17% |
| Mi | 370 | 10.1% | 322 | 10.52% | 1288 | 23.91% |
| Mo | 999 | 27.28% | 874 | 28.56% | 874 | 16.22% |
| Se | 193 | 5.27% | 180 | 5.89% | 720 | 13.36% |
| Pr | 295 | 8.05% | 274 | 8.95% | 1096 | 20.34% |
| **Total** | **3662** | - | **3060** | - | **5388** | - |

### F. Tone Mapping Images

Tone mapping a digital image processing technique to convert a High Dynamic Range (HDR) image to a Low Dynamic Range (LDR) image. The conversion is particularly useful since the latter version is more suitable to display. Tone mapping is done by compressing the dynamic range of the HDR image while preserving the details of the image. Each tone in the original image, which has a broad dynamic range, is mapped to achieve a predetermined and viewable tone. Tone mapping is performed using an operator $\mathcal{T}$, also known as the Tone Mapping Operator (TMO), which is defined in general as [10]:

$$\mathcal{T}\{I\} = \begin{cases} L_d = \mathcal{T}_L\{H_d\} &: \mathbf{R}_i^{h \times w} \\ \begin{bmatrix} R_d \\ G_d \\ B_d \end{bmatrix} = L_d \left(\frac{1}{L_w}\begin{matrix} R_w \\ G_w \\ B_w \end{matrix}\right)^s \end{cases} \quad (1)$$

where $I$ is the original image, $h$ and $w$ are its height and width respectively, $L_d$ and $H_d$ are LDR and HDR luminance value of a pixel respectively, $\mathbf{R}_i \subseteq \mathbf{R}$, and $s \in (0,1]$ is the saturation factor. In this study, we used the global tone mapping technique where the mapping is applied to all the pixels with the same operator ($\mathcal{T}$). However, $\mathcal{T}$ can be applied to the pixels in many ways. The common forms are linear scaling, logarithmic mapping, and exponential mapping [10]. In the first method, the original image is simply multiplied by a predefined scaling factor $e$ such that:

$$L_d\{p_i\} = eH_d(p_i) \quad (2)$$

where, $p_i$ is a sample pixel. Logarithmic mapping uses a logarithmic function to calculate the LDR luminance values. However, we used exponential mapping that uses an exponential function to calculate the luminance value of each pixel of the image. The relationships between $L_d$ and $H_d$ for this mapping are defined in eq. (3) and (4) respectively:

$$L_d(p_i) = \frac{\log_{10}(1 + qH_d(p_i))}{\log_{10}(1 + k \times \max(H_d(p_i)))} \quad (3)$$

$$L_d(p_i) = 1 - \exp(-\frac{H_d(p_i)}{k \times \text{mean}(H_d(p_i))}) \quad (4)$$

where $q \in (1, \infty]$ and $k \in [\infty, 1)$ are user-defined constants that determine the output of the mapping operation. Fig. 5 displays a sample retinal image and its global tone-mapped version which reveals information that is more suitable for classification than the original image.

*G. Omitting Black Corners and Calculating Histogram*

The histogram simply refers to the frequency of each distinct intensity value of pixel present in the image. Each pixel of an RGB image contains three intensity values (between 0 and 255) in three channels. An exhaustive search over all the pixels reveals the number of occurrences of each value in all three channels, which in turn will portray the image's histogram. Mathematically, for a single channel:

$$H(l) = \sum_{i=1}^{I'_h} \sum_{j=1}^{I'_w} k \quad (5)$$

where $\quad k = \begin{cases} 1, & \text{if } I'(i,j) = l \\ 0, & \text{otherwise} \end{cases}$

$I'_h$ and $I'_w$ are the height and width of the image, respectively. However, in our case, most of the cropped images contain different amounts of black pixels at the corners (Fig. 5(c)), which do not carry any significant information and need to be removed before calculating the histogram. *Algorithm-II* narrates the entire process of tone mapping, black corner detection and omission, and histogram calculation. Fig. 6 illustrates the distribution of the samples in a two-dimensional plane based on the three-channel color intensity information of the retinal images.

| *Algorithm-II:* | | |
|---|---|---|
| **Input** | : | An image with unwanted black corners, $I'$ |
| **Output** | : | Flatten histogram of the image without unwanted black corners, $H'$ |
| *Step-1* | : | Create a $n \times 2$ matrix ($M$) to store the indexes of the pure black pixels. |
| *Step-2* | : | **for** $i = 1$ to $I'_h$<br>  **for** $j = 1$ to $I'_w$<br>    **if** the $(i,j)$ pixel of $I'$ contains zero in all the three channels (r, g, b)<br>      store the index value pair $(i,j)$ to $M$<br>  **end all for** |
| *Step-3* | : | Calculate the tone map of $I'$ and store it to $I'_T$ [Using eq. (4)] |
| *Step-4* | : | Create an all-zero $3 \times 256$ matrix ($H$) |
| *Step-5* | : | **for** $i = 1$ to $I'_h$<br>  **for** $j = 1$ to $I'_w$<br>    **if** index value pair $(i,j)$ is not present in $M$<br>      **for** $k = 0$ to $255$<br>        **for** $l = 1$ to $3$<br>          **if** $I'_T(i,j,l)$ has the same value as $k$<br>            add 1 to $H(l, k+1)$<br>  **end all for** |
| *Step-6* | : | Concatenate the rows of $H$ and save to $H'$ |
| *Step-7* | : | **return** $H'$ |

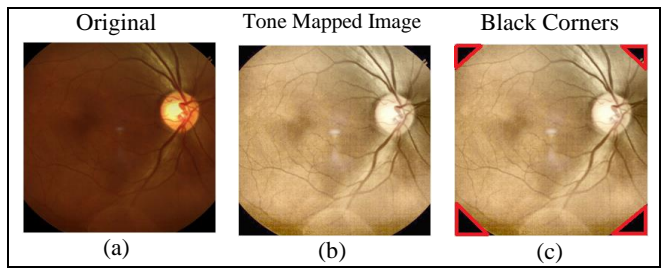

Fig. 5. (a) A sample retinal image, (b) its tone-mapped view, and (c) areas that do not contain necessary information.

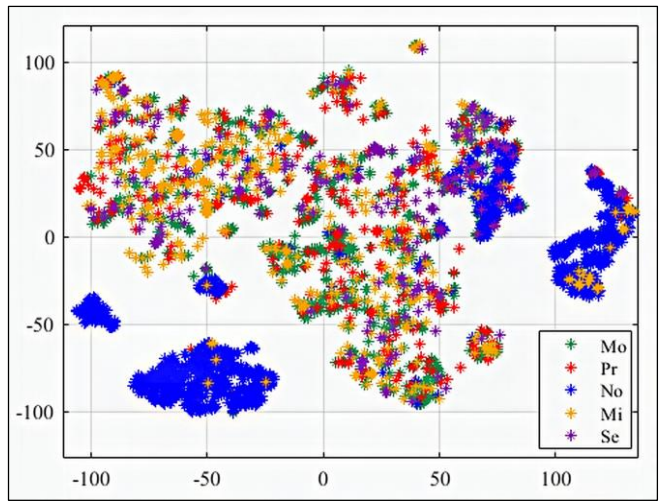

Fig. 6. t-SNE of the operational dataset.

*H. Classification of Retinal Images Using ET classifier*

ET classifier is an ensemble learning technique based on decision trees and bagging learning techniques used to solve classification and regression problems. The classifier increases the randomization while splitting the nodes of a decision tree, which in turn, reduces the training time and variances in decisions, and improves the classification accuracy. This supervised learning method was proposed by Pierre Geurts in a paper published in 2006 [11]. Each tree within the ET classifier partitions the training dataset into multiple disjoint sections. Each decision tree then provides a constant prediction from each section by taking the mean value of the outputs of the samples of that section. Considering a dataset containing $N$ training samples, the sections can be expressed as:

$$\mathcal{S} = \{(x_i, y_i) \mid i = 1, 2, \ldots, N\} \quad (6)$$

where $x_i$ is the $i$th input sample and $y_i$ is the corresponding output. Now, if we consider a decision tree with index $i_t$ within the ensemble and a function $f_{i_t}(x)$ that associates input $x$ to its corresponding region in the partition, the approximate output can be calculated using [12]:

$$\mathsf{y}_{i_t} = \sum_{i=1}^{N} \mathcal{K}(x, x_i) y_i \quad (7)$$

where $\mathcal{K}$ is a function defined by:

$$\mathcal{K}(x, x_i) = \frac{\mathcal{J}(x_i \in f_{i_t}(x))}{\sum_{i'=1}^{N} \mathcal{J}(x_{i'} \in f_{i_t}(x))} \quad (8)$$

where $\mathcal{J}$ is the indicator function and defined as:

$$\mathcal{J} = \begin{cases} 1, & argument = true \\ 0, & argument = false \end{cases}$$

The final approximation is an average of the decisions of $N_t$ trees, which can be calculated using:

$$y(x) = \frac{1}{N_t}\sum_{i_t=1}^{N_t} y_{i_t}(x) \quad (9)$$

The final outcome is determined using the kernel function $\mathcal{K}$ by the following equation derived from eq. (7) and (8):

$$\mathcal{K}(x, x_i) = \frac{1}{N_t}\sum_{i_t=1}^{N_t} \frac{\mathcal{J}(x_i \in f_{i_t}(x))}{\sum_{i'=1}^{N} \mathcal{J}(x_{i'} \in f_{i_t}(x))} \quad (10)$$

The ET algorithm can use any objective function to optimize splits of the nodes of the tree. However, in classification problems, usually, a normalized version of the information gain-based decision tree structure is used [13].

### III. RESULTS AND DISCUSSION

In the previous sections, we discussed the retinal images of the APTOS 2019 Blindness Detection dataset, the challenges to work with these images, and how we planned to overcome them. In this section, we will present the outcomes of our classification model. As mentioned earlier, we have used an ensemble learning method, ET classifier, to classify different levels of blindness. Before we performed the classification, we needed to tune a few parameters of the classifier itself to achieve the optimum performance. More specifically, we tuned the parameters called "n_estimators" and "max_features." The former one simply determines the number of decision trees in the forest, which can be any integer number, and the latter determines how many features the classifier will consider while looking for the best split. Fig. 7 presents the Out-of-bag (OOB) error while working with the provided dataset at various values of n_estimators and different setup of max_features. OOB error is used to measure the prediction error of bagging models.

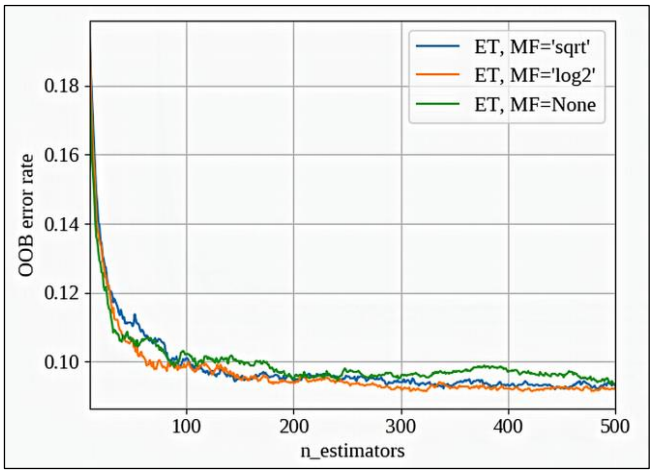

Fig. 7. OOB error rates of the ET classifier at various settings.

The figure depicts that the error rate decreases as the number of estimator is increased. However, choosing a very high value for this parameter will awfully slow down the training process [14]. Looking at the graph, the error rate is considerably low for all three max_features conditions when is n_estimator is 200. Since the "log2" curve has the lowest amplitude at 200, we chose to work with 200 and "log2" as the tuned values of those two parameters, respectively, for the rest of the experiment.

Now that we have the ET classifier optimized to provide the maximum performance on the prepared dataset, we used 70% of the images (i.e., their corresponding features) to train the algorithm, and the rest 30% images to test it. However, the result of a single classification may not be a plausible determiner, especially when the data are this noisy, and there is a possibility for the model to get biased. So, we performed 10-fold cross-validation so that each sample is guaranteed to belong in the training subset and the test subset for at least once. We present the iteration-wise classification accuracies in Fig. 8. The accuracy scores range from 87.67% to 92.87%. However, according to the convention, the algebraic mean score of all the iterations is to be taken as the true classification accuracy of the model. In our case, that score is 91.07%; the level has been marked with a red line in the figure for easy interpretation.

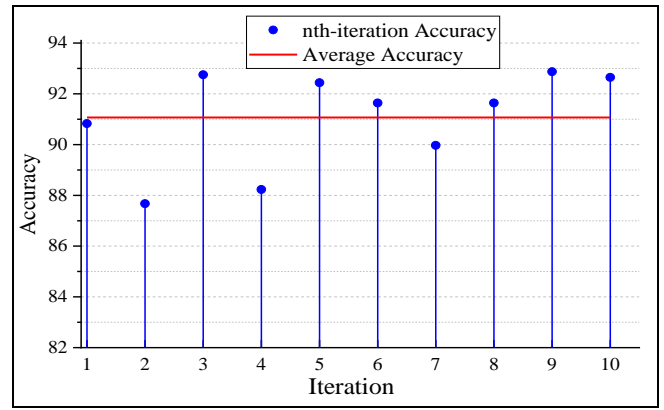

Fig. 8. Accuracies at different iterations of the 10-fold cross-validation.

Another credible way to judge the performance of a classifier is to take a look at the confusion matrix of its classification outcome. The confusion matrix puts the predicted labels of the test samples against the actual labels of those samples in a matrix-like fashion and helps us to visualize how many samples of each class were classified correctly, and how many were not along with the erroneous output classes. If the classification outcomes a perfect accuracy, only the diagonal cells (top-right to bottom-left) would contain integer numbers, and all the other boxes would contain zeros [15]. For the classification results that are less than the ideal, single or multiple non-diagonal boxes will contain the number of misclassified samples corresponding to the real and predicted labels. Fig. 9 provides the confusion matrix of our classification. The imparted matrix is an algebraic sum of all the ten matrices acquired from the 10-fold cross-validation. The matrix shows that images that belong to the "Moderate DR" class were misclassified the most. The extent of misclassification of the other four classes was somewhat acceptable, considering the fact that they are an aggregation of 10 different classification operations.

Some other parameters can also be derived from Fig. 9, which would make the potency of the proposed model even more apparent. We can calculate the class-wise accuracy, which is known as Recall; the true-positive rate, also known as Precision; and a weighted average of the former two that explains the performance of a classifier better than the accuracy score, known as F1-score. Table III presents these scores of the described ET-based classification model.

Fig. 9. Aggregated confusion matrix of the classifications.

The table shows that the class labeled as "Mo" has scored the least in all three parameters, which reflects the discussion in the previous paragraph. However, the other four classes have scored over 90% in all criteria meaning the classifier was highly accurate in predicting the labels of the samples that belong to those classes.

TABLE III. PRECISION, RECALL, AND F1-SCORES

| Class | Precision (%) | Recall (%) | F1-score (%) |
|---|---|---|---|
| No | 92.92 | 96.07 | 94.47 |
| Mi | 89.96 | 96.77 | 93.24 |
| Mo | 81.35 | 63.95 | 71.62 |
| Se | 93.85 | 95.71 | 94.77 |
| Pr | 93.91 | 95.18 | 94.55 |
| **Average** | **90.4** | **89.54** | **89.97** |

Furthermore, we present the Receiver Operating Characteristics (ROC) curves of each class in Fig. 10. The ROC curve of a class demonstrates the performance of the classifier for all values of the discrimination threshold. Basically, it is a plot of the classifier's recall scores against its false-positive rates [16]. Unlike accuracy values, the ROC curves are not affected by the imbalanced proportion of class samples in the operational dataset, and hence, provide a more unbiased view of the performance of the classifier. Quite expectedly, we can see the "Mo" curve is situated farthest away from the top-right corner emphasizing the struggle of the classifier to recognize the mentioned class.

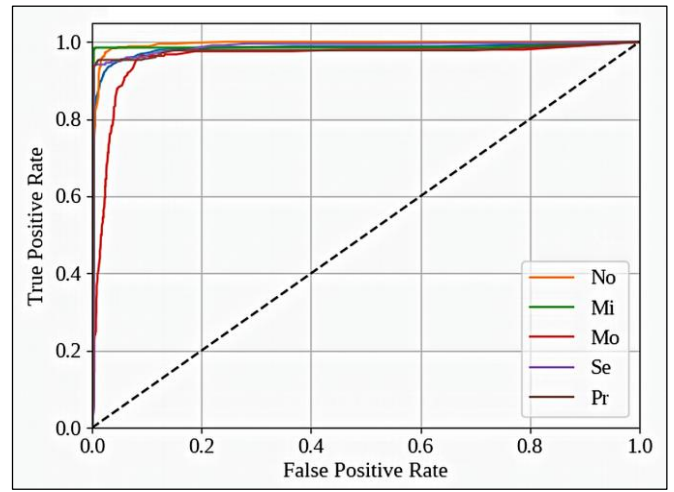

Fig. 10. ROC curves of each class.

IV. CONCLUSIONS

The paper described a blindness detection method designed using an ensemble ML algorithm that analyzes retinal images to determine the label of blindness. Images used in this study are noisy and mislabeled and were collected from a Kaggle competition [9]. After eliminating the noisy samples manually and augmenting the samples of certain classes to balance the dataset, ET classifier was put into task based on the color information of the images. The model achieved a 91% classification accuracy on average, which makes it very useful in detecting the severity of blindness. However, the performance of the model in identifying some classes is still far less than the optimum, which leaves a vast scope to work with it in the future. In truth, the ingenuousness of the dataset makes it particularly interesting to work with, even though it contains many defective samples, and reports suggest improper labeling and duplication of retinal images. Nonetheless, we are also eager to work with the images that were excluded due to their poor conditions in our future studies.

REFERENCES

[1]  F. Bandello, M. A. Zarbin, R. Lattanzio, I. Zucchiatti, and Springer-Verlag GmbH, *Clinical Strategies in the Management of Diabetic Retinopathy A step-by-step Guide for Ophthalmologists*. 2014.

[2]  B. Lumbroso, M. (Ophthalmologist) Rispoli, and M. C. Savastano, *Diabetic retinopathy*. .

[3]  J. Chua, C. X. Y. Lim, T. Y. Wong, and C. Sabanayagam, "Diabetic retinopathy in the Asia-pacific," *Asia-Pacific Journal of Ophthalmology*, vol. 7, no. 1. Asia-Pacific Academy of Ophthalmology, pp. 3–16, 01-Jan-2018.

[4]  A. Akhter, K. Fatema, S. F. Ahmed, A. Afroz, L. Ali, and A. Hussain, "Prevalence and associated risk indicators of retinopathy in a rural Bangladeshi population with and without diabetes.," *Ophthalmic Epidemiol.*, vol. 20, no. 4, pp. 220–7, Aug. 2013.

[5]  M. U. Akram, S. Khalid, and S. A. Khan, "Identification and classification of microaneurysms for early detection of diabetic retinopathy," *Pattern Recognit.*, vol. 46, no. 1, pp. 107–116, Jan. 2013.

[6]  B. Antal and A. Hajdu, "An ensemble-based system

for automatic screening of diabetic retinopathy," *Knowledge-Based Syst.*, vol. 60, pp. 20–27, 2014.

[7] S. Wang, Y. Yin, G. Cao, B. Wei, Y. Zheng, and G. Yang, "Hierarchical retinal blood vessel segmentation based on feature and ensemble learning," *Neurocomputing*, vol. 149, no. PB, pp. 708–717, Feb. 2015.

[8] E. Saleh *et al.*, "Learning ensemble classifiers for diabetic retinopathy assessment," *Artif. Intell. Med.*, vol. 85, pp. 50–63, Apr. 2018.

[9] "APTOS 2019 Blindness Detection," 2019. [Online]. Available: https://www.kaggle.com/c/aptos2019-blindness-detection/. [Accessed: 10-Jun-2019].

[10] F. Banterle, A. Artusi, K. Debattista, and A. Chalmers, *Advanced high dynamic range imaging*. 2017.

[11] P. Geurts, D. Ernst, and L. Wehenkel, "Extremely randomized trees," *Mach. Learn.*, vol. 63, no. 1, pp. 3–42, Apr. 2006.

[12] L. Busoniu, R. Babuska, B. De Schutter, and D. Ernst, *Reinforcement learning and dynamic programming using function approximators*. 2010.

[13] A. Criminisi and J. Shotton, *Decision Forests for Computer Vision and Medical Image Analysis*. Springer, 2013.

[14] M. Ben Fraj, "In Depth: Parameter tuning for Random Forest," 2017. [Online]. Available: https://medium.com/all-things-ai/in-depth-parameter-tuning-for-random-forest-d67bb7e920d. [Accessed: 01-Aug-2019].

[15] N. Sikder, M. S. Chowdhury, A. M. Shamim Arif, and A.-A. Nahid, "Human Activity Recognition Using Multichannel Convolutional Neural Network," *2019 5th Int. Conf. Adv. Electr. Eng.*, 2019.

[16] G. Hackeling, *Mastering Machine Learning with scikit-learn*. Packt Publishing, 2014.